\documentclass[electronic]{vgtc}             

\ifpdf
  \pdfoutput=1\relax                   
  \usepackage{graphicx}                
  \DeclareGraphicsExtensions{.pdf,.png,.jpg,.jpeg} 
\else
  \usepackage{graphicx}                
  \DeclareGraphicsExtensions{.eps}     
\fi%

\graphicspath{{figures/}{pictures/}{images/}{./}} 

\usepackage{microtype}                 
\PassOptionsToPackage{warn}{textcomp}  
\usepackage{textcomp}                  
\usepackage{mathptmx}                  
\usepackage{times}                     
\usepackage{cite}                      
\usepackage{tabu}                      
\usepackage{booktabs}                  
\usepackage[frozencache,cachedir=.]{minted}
\usepackage{listings}

\onlineid{1114}

\vgtccategory{Research}

\vgtcinsertpkg

\title{Layout Aware Inpainting for Automated Furniture Removal in Indoor Scenes}

\author{Prakhar Kulshreshtha\thanks{e-mail: prakhar@geomagical.com}\\ %
        \scriptsize Geomagical Labs, Inc. %
\and Konstantinos-Nektarios Lianos\thanks{e-mail: nelianos@geomagical.com}\\ %
     \scriptsize Geomagical Labs, Inc. %
\and Brian Pugh\thanks{e-mail: bpugh@geomagical.com}\\ %
     \scriptsize Geomagical Labs, Inc. %
\and Salma Jiddi\thanks{e-mail: salma@geomagical.com}\\ %
     \scriptsize Geomagical Labs, Inc.}

\teaser{
  \centering
  \includegraphics[width=\linewidth]{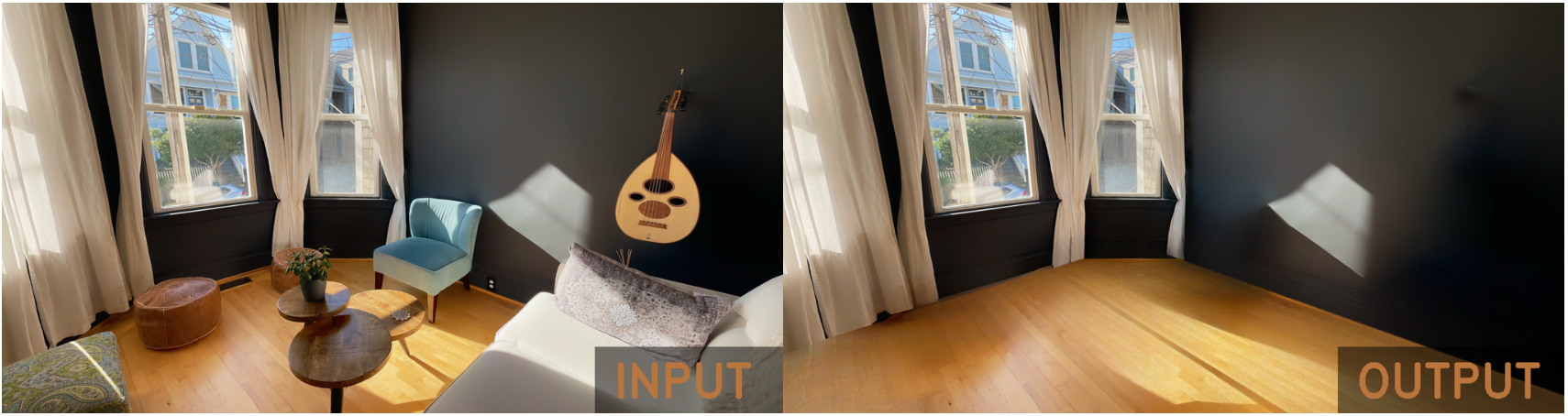}
  \caption{Furniture Removal Results. Left is the input image with furniture. Right is our predicted empty room.}
  \label{fig:teaser}
}

\abstract{
We address the problem of detecting and erasing furniture from a wide angle photograph of a room. Inpainting large regions of an indoor scene often results in geometric inconsistencies of background elements within the inpaint mask. To address this problem, we utilize perceptual information (e.g. instance segmentation, and room layout) to produce a geometrically consistent empty version of a room. We share important details to make this system viable, such as per-plane inpainting, automatic rectification, and texture refinement. We provide detailed ablation along with qualitative examples, justifying our design choices. We show an application of our system by removing real furniture from a room and redecorating it with virtual furniture.
} 

\CCScatlist{
  \CCScatTwelve{Augmented Reality}{Diminished Reality}{Computer vision}{Inpainting}
}

\renewcommand{\listingscaption}{Algorithm}

\begin{document}

\firstsection{Introduction}
\label{sec:intro}
\maketitle

The ability to remove objects from a scene is a common task in applications like image editing, Augmented Reality, and Diminished Reality \cite{bardi_2016, shoei_2017}.
The general problem of image inpainting has seen many improvement over the past few decades in both classical and deep learning approaches \cite{patchmatch, deepfillv2, comodgan, lama}.
Modern inpainting techniques work incredibly well for small to medium sized regions \cite{lama}, but still struggle to produce convincing results for larger missing segments.
For these regions, the texture and structure from surrounding areas fail to propagate in a visually pleasing, and physically plausible way.
Inpainting large regions requires geometric, texture, and lighting consistency to produce convincing results.
State-of-the-art inpainting networks like \cite{lama} often fail to complete large global structures, like the plausible continuation of walls, ceilings, and floors in an indoor scene.  

In this work, we address the challenges of inpainting large regions of an indoor scene, and propose a system for synthesizing a view of the room with all objects removed.
To do this, we introduce multiple novel steps that are not present in generic inpainting systems.
The contribution of this work is as follows:
\begin{itemize}
    \itemsep0em 
    \item A strategy for applying inpainting networks trained on in-the-wild images on rectified planes. 
    \item Refining the inpainted texture using a combination of offline training and run-time optimization.
    \item An end-to-end system for automatically detecting and erasing furniture from a room, and its application as an interior design product (\autoref{fig:teaser}).
\end{itemize}

\section{Related Work}
\label{sec:related}
\begin{figure*}[h!]
 \centering 
 \includegraphics[width=\textwidth]{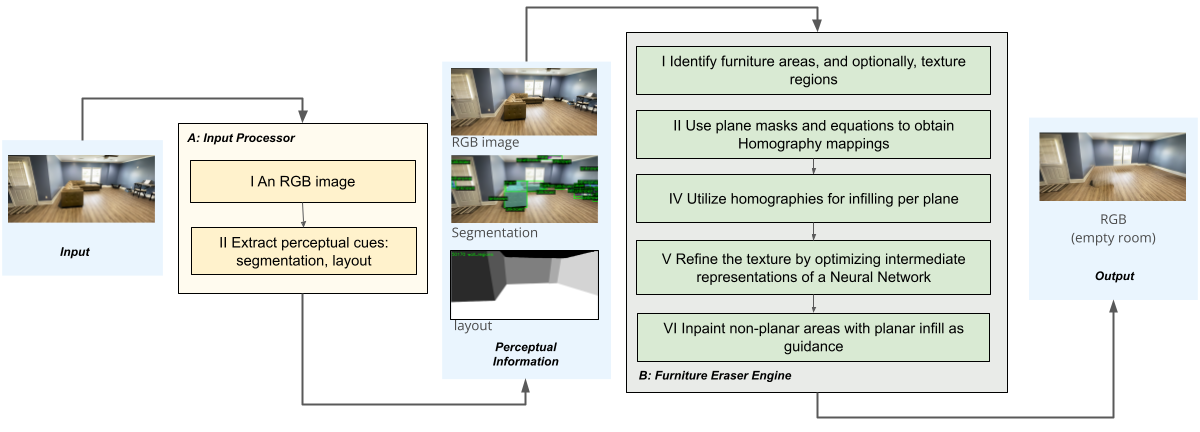}
 \caption{System Overview. Input data is fed through an Input Processor (A) to generate intermediate perceptual artifacts like segmentation map and room layout. This perceptual data is fed into the furniture eraser engine (B) where image inpainting is performed on each independent rectified plane in the scene. The results are blended together to obtain an image of the empty room.}
 \label{fig:system}
\end{figure*}

\begin{figure}[h!]
    \centering
    \includegraphics[width=\columnwidth]{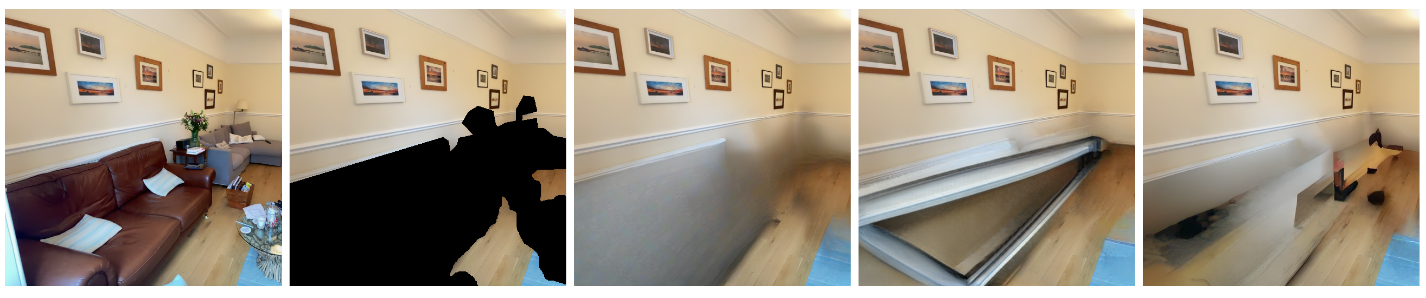}
    \caption{From left to right: input image, masked image, and then inpainting results using \cite{lama}, \cite{comodgan} and \cite{latentdiffusion} respectively. The results are blurry, color inconsistent, and hallucinate new unwanted structure.}
    \label{fig:trying_dnns}
\end{figure}

A number of approaches have been proposed for solving image inpainting, including color-diffusion-based methods \cite{telea, bartelmio}, patch-based techniques \cite{patchmatch, exemplarbased}, convolutional neural networks \cite{partialconv, deepfillv1, deepfillv2, edgeconnect, lama}, and diffusion models \cite{latentdiffusion}.
Deep Neural Networks (DNN) have been observed to perform well at inpainting small and medium-sized holes \cite{lama, comodgan} because of their ability to encode both local and global context.
Improving the receptive field of a neural network to improve performance has been the focus of developments like fourier convolutions\cite{lama}, transformers\cite{mat}, and diffusion models\cite{latentdiffusion}.
Despite these improvements, these networks still struggle to inpaint an indoor scene with large unknown regions (\autoref{fig:trying_dnns}).

Inpainting can be constrained using edges\cite{edgeconnect}, semantic segmentation\cite{ntavelis2020sesame}, and patches from known regions\cite{neuralpatch}.
\cite{huang2014image} utilizes planar information to produce perspective corrected patches prior to applying patched-based inpainting.
\cite{kawai2015diminished} performs patch-based per-plane inpainting, but it requires user input to determine the resolution of the rectified image.
\cite{gkitsas2021panodr} utilizes room layout information to empty a spherical panorama image of a furnished room.
Their system was trained end-to-end on layout and empty room data, which are difficult to collect.
\cite{zhang2021no} incorporates lighting and geometry constraints by coupling an intrinsic image decomposition network, a differentiable shadow removal module, and an inpainting network.

Directly trying to solve the same problem, \cite{zhang2016emptying} also estimates an empty version of a room.
Their system recovers the light sources and the surface reflectance of various elements of the scene, but it has been shown to only work on rooms with relatively simple texture and geometry. 
A completely different approach to indoor inpainting is to estimate the room with a CAD model\cite{izadinia2017im2cad}.
While IM2CAD maintains the layout of the room, it replaces floors with generic wood texture, and walls with a median color that may not accurately represent the scene.


In our system, we constrain inpainting of an indoor scene by exploiting the 3D planar layout of a room.
Similar to \cite{kawai2015diminished}, we perform inpainting on a per-rectified-plane basis.
However, we use a DNN instead of a patch-based algorithm to perform the actual inpainting.
Unlike \cite{kawai2015diminished}, we do not need to explicitly set the rectification resolution because it is tied to the training resolution of the inpainting network.
Our system is modular, consisting of several independent neural networks and classical computer vision algorithms, allowing each component to be independently updated as new improvements are developed in common computer vision tasks.
Our inpainting network is trained on in-the-wild indoor and outdoor images, without needing any manual annotations.

\section{System Overview}
\label{sec:system}

Our system is composed of an input processor and a furniture eraser engine (see \autoref{fig:system}).
The input processor parses the input image using existing methods to obtain perceptual cues such as segmentation masks \cite{resnest, beit, queryinst, mask2former} and the room layout \cite{liu2019planercnn}.
While we recognize that extracting any of these perceptual cues is not a trivial task, in this work, we assume these cues to be a part of the input, and explicitly focus on the inpainting part of our system.
In the subsequent subsections, we describe how the furniture eraser engine utilize these cues for inpainting.

\subsection{Inpainting mask}
\begin{figure}[h]
    \centering
    \includegraphics[width=\columnwidth]{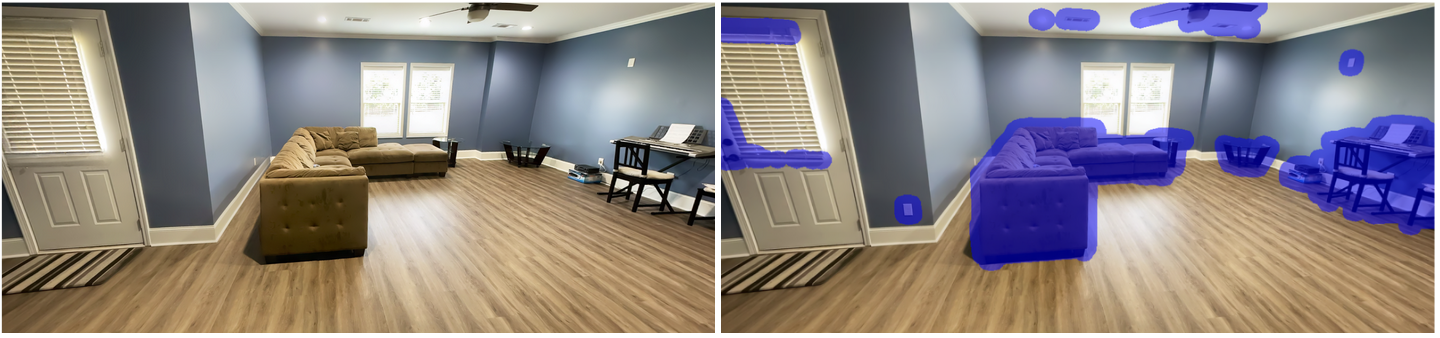}
    \caption{Utilizing the semantic information, we extract an inpainting mask (blue regions) comprising of all the objects in the scene.}
    \label{fig:fmask}
\end{figure}

The furniture eraser engine uses instance segmentation to identify all the objects in the scene.
The union of all the object masks becomes the inpainting mask for our task of emptying the room (see Fig. \ref{fig:fmask}).
Our goal is to replace all the pixels in this mask with the background texture.

\subsection{Plane-wise inpainting}
\begin{figure}[h]
    \centering
    \includegraphics[width=\columnwidth]{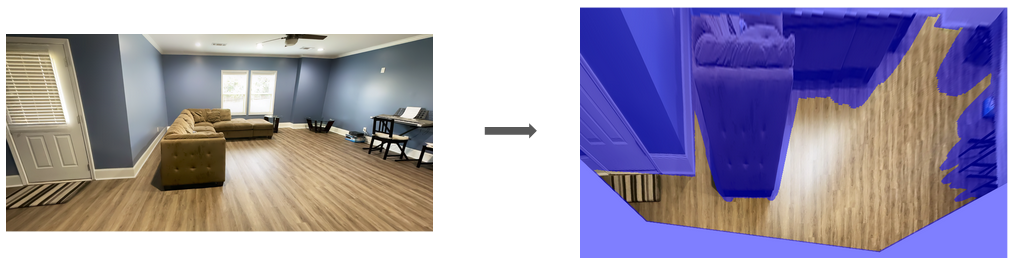}
    \caption{Rectification of the floor plane before inpainting. The blue areas in the second image show the inpainting mask to the neural network.}
    \label{fig:rectification}
\end{figure}

We observe that inpainting an indoor scene suffers from two major problems: perspective distortion, and context mixing from different texture regions of an image.
These problems are especially pronounced when we remove large objects from a scene \cite{lama, comodgan}.
We utilize the room layout information to solve this problem.
Each wall and floor of the room can be represented as a plane in 3D space.
Given the 3D equation of a plane, along with its binary occupancy mask in the image, we can rectify the plane to become fronto-parallel to the camera (\autoref{fig:rectification}).
We determine the resolution of the rectified plane based on the native training resolution of the inpainting network, since a neural inpainting network achieves optimal performance around the resolution at which it was originally trained\cite{featurerefinement}.  

\subsection{Utilizing a pretrained DNN}

In this section, we discuss how to utilize a Neural Network pretrained on the task of image inpainting in the wild, for inpainting indoor scenes.
As shown in third row in \autoref{fig:abl}, directly using a network on simple perspective images results in a very blurry output.
This happens because the receptive field of the network is limited, which makes it challenging to complete the geometry and semantic regions in large holes.
To overcome these issues, we inpaint each rectified plane separately. The inpaint mask is defined to be the union of furniture mask and the unknown area (since rectification would introduce some out-of-frame pixels in the rectified view).
This image-mask pair is input to the inpainting network, and then output is unrectified to fill in the missing pixels of that plane.
Per-plane inpainting substantially improves the inpainting quality (fourth row of \autoref{fig:abl}). 


\subsection{Texture refinement}

We observe that inpainting networks trained on the Places2\cite{places} dataset often infills large regions with an undesired gray texture (See Fig. \ref{fig:abl}).
To remedy this, we prepare a more diverse mixture of around 1 million RGB images, from SUN-RGBD\cite{zhou2014learning, silberman2012indoor, janoch2013category, xiao2013sun3d}, Diode\cite{diode_dataset}, Unsplash\cite{unsplash}, Hypersim\cite{roberts:2021}, OpenImages\cite{openimages}, Google-Landmark\cite{weyand2020GLDv2}, and in-house collected RGB images of real and synthetic rooms.
We trained on the entire dataset for first 600k iterations, then we transferred the weights from first session, and trained on only the indoor scenes for another 300k iterations. 

\begin{figure}[h]
    \centering
    \includegraphics[width=0.6\columnwidth]{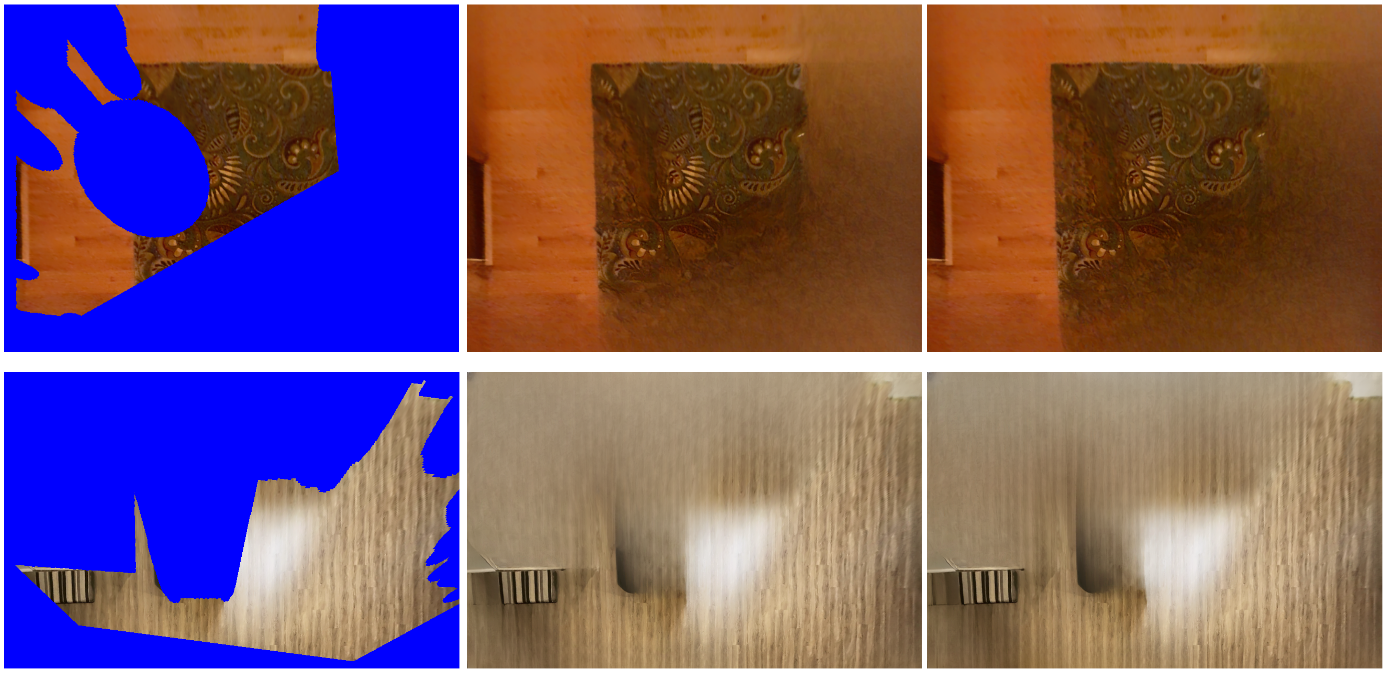}
    \caption{Two examples of feature refinement. For each row, left-most image is the masked input. Middle image is the prediction using \cite{lama} trained on our dataset. Rightmost image is feature-refinement applied to the output shown in the middle image. The carpet structure and the wooden textures are greatly improved.}
    \label{fig:refinement}
\end{figure}

In addition to retraining the neural network, we also apply the featuremap refinement approach described in \cite{featurerefinement}.
We use the multi-scale loss of \cite{featurerefinement}, and add a color histogram loss \cite{afifi2021histogan} between the histogram of the unmasked pixels and the histogram of the inpainted image at each scale.
This histogram loss helps in getting rid of the gray areas while inpainting very large holes in relatively homogeneous images (\autoref{fig:refinement}).

These two modifications significantly refine the quality of the infilled texture, when compared to directly using off-the-shelf model of \cite{lama}. 

\subsection{Inpainting non-planar areas}

\begin{figure}[h]
    \centering
    \includegraphics[width=0.8\columnwidth]{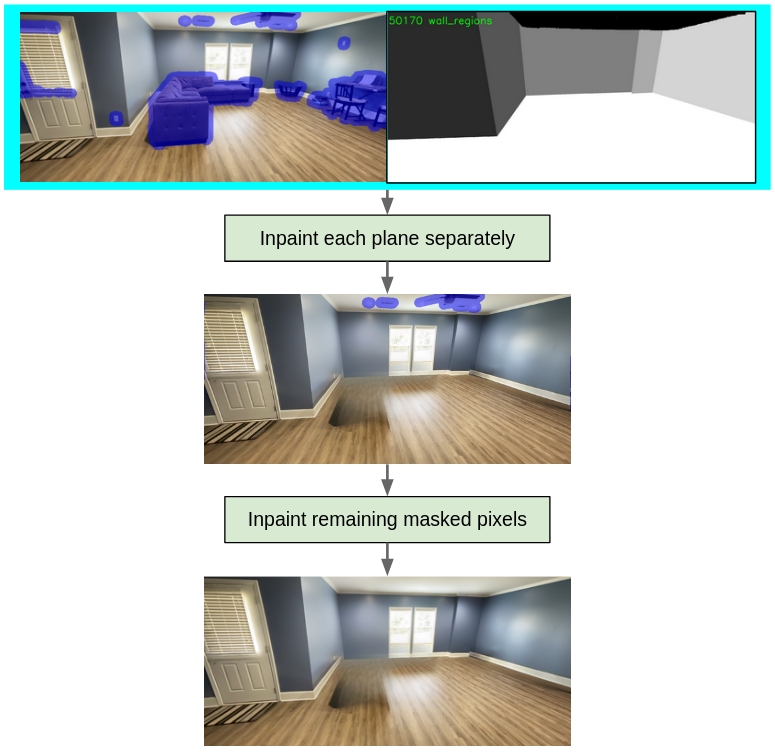}
    \caption{Inpainting of non-planar regions. The planar regions are inpainted first, and then we do a final forward pass to inpaint the remaining masked areas (observe the ceiling fan and ceiling light are inpainted in the bottom image).}
    \label{fig:remaining}
\end{figure}

Since we are inpainting in a per-plane fashion, some pixels, which aren't a part of any plane, will be left out.
To inpaint these regions, we first replace all the pixels belonging to planar masked regions with their inpainted values.
This image along with the remaining mask is input to the inpainting network.
The output to this forward pass is the final inpainted image (\autoref{fig:remaining}).

\section{Evaluation}
\label{sec:evaluation}

In this section, we discuss how we prepare the evaluation testset, followed by the metrics we chose.
We perform an ablation study to identify the contributions of each component in our system.

\subsection{Test Set}

To evaluate our system, we collect a set of 229 captures of empty room scenes.
For each scene, we prepare inpainting mask by using silhouettes of virtually placed furniture.
The ground truth is set to be the original unmasked image.
The inpainting system is then tasked to inpaint this missing region.
A limitation of this method, as mentioned in \cite{gkitsas2021panodr}, is that virtual masks in empty room images do not cast shadows, and do not change the lighting of the room by blocking a light source such as a window or a lamp.
Despite that, it's a useful test-set to evaluate the inpainting performance, and perhaps the best ground-truth one could get for natural indoor scenes.

\subsection{Metrics}

Earlier literature used to evaluate the inpainting performance on pixelwise metrics, like PSNR, or mean squared error \cite{edgeconnect, partialconv}, but these metrics aren't suitable to evaluate the performance on large mask inpainting, where there can be a variety of visually distinct, but plausible solutions.
Recent work\cite{lama, zits} has shifted to perceptual metrics, like Learnt Perceptual Similarity Score (LPIPS)\cite{lpips} and FID\cite{fid}.
Our set of scenes is just 229, so we do not calculate FID score because on a very small dataset, its estimated value significantly differs from its true value\cite{chong2020effectively}.
For these reasons, our primary metric of comparison will be LPIPS.
Despite its limitations, we also report PSNR for completeness.

In addition to this, we found that the visual experience of an empty room substantially deteriorates if the inpainting introduces discontinuities or edges.
To quantify this, we introduce a new metric - incoherence. To calculate incoherence, we first extract the edge-probability map \cite{xie2015holistically, pytorch-hed} for both the ground-truth and the predicted image. All the pixels in predicted image, for which the corresponding pixel is an edge in the ground-truth image, are suppressed to $0$. Incoherence is then the average of edge probabilities across all the pixels in the predicted image. A higher incoherence can therefore be associated with more, or stronger, false edges in the inpainting. 
The pseudo-code for this metric is given in \autoref{alg:incoherence}.

\begin{listing}[!htb]
\begin{minted}[fontsize=\scriptsize]{python}

def calc_incoherence(image_gt, image_pred, inpaint_mask):
    edge_gt = gaussian_blur(edgemap(image_gt))
    edge_pred = edgemap(image_pred)
    # enhance all the edges above 0.1 probability
    edge_gt[edge_gt>0.1] = 1.0 
    imask = edge_pred - edge_gt
    imask[imask <= 0.01] = 0.0
    incoherence = np.mean(imask[inpaint_mask]) 
    return incoherence

\end{minted}
\caption{Python pseudocode for calculating incoherence.}
\label{alg:incoherence}
\end{listing}

\begin{figure}[h]
    \centering
    \includegraphics[width=\columnwidth]{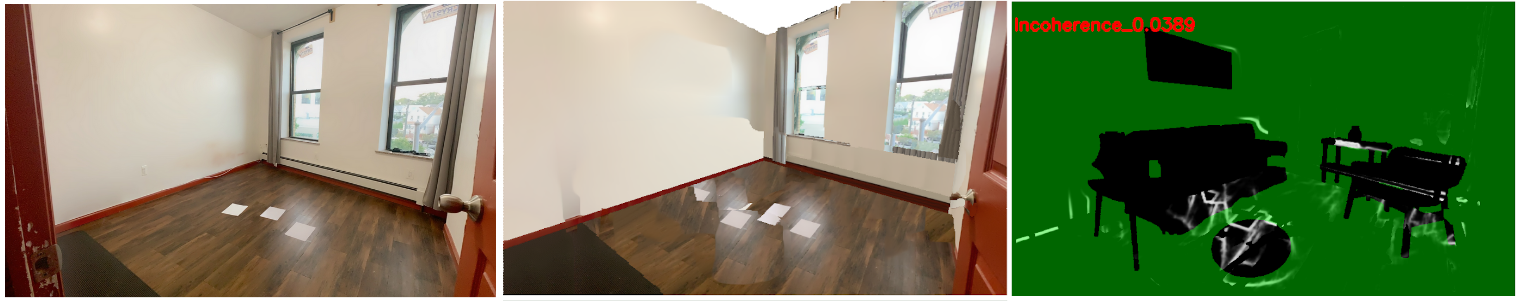}
    \caption{Incoherence visualization. From left to right: ground-truth image, inpainted image, incoherence map. The green/black colors are for unmasked/masked regions, and the white color shows the incoherence introduced by the inpainting.}
    \label{fig:incoherence}
\end{figure}

\subsection{Results and Discussion}

\begin{figure*}
    \centering
    \includegraphics[width=0.8\textwidth]{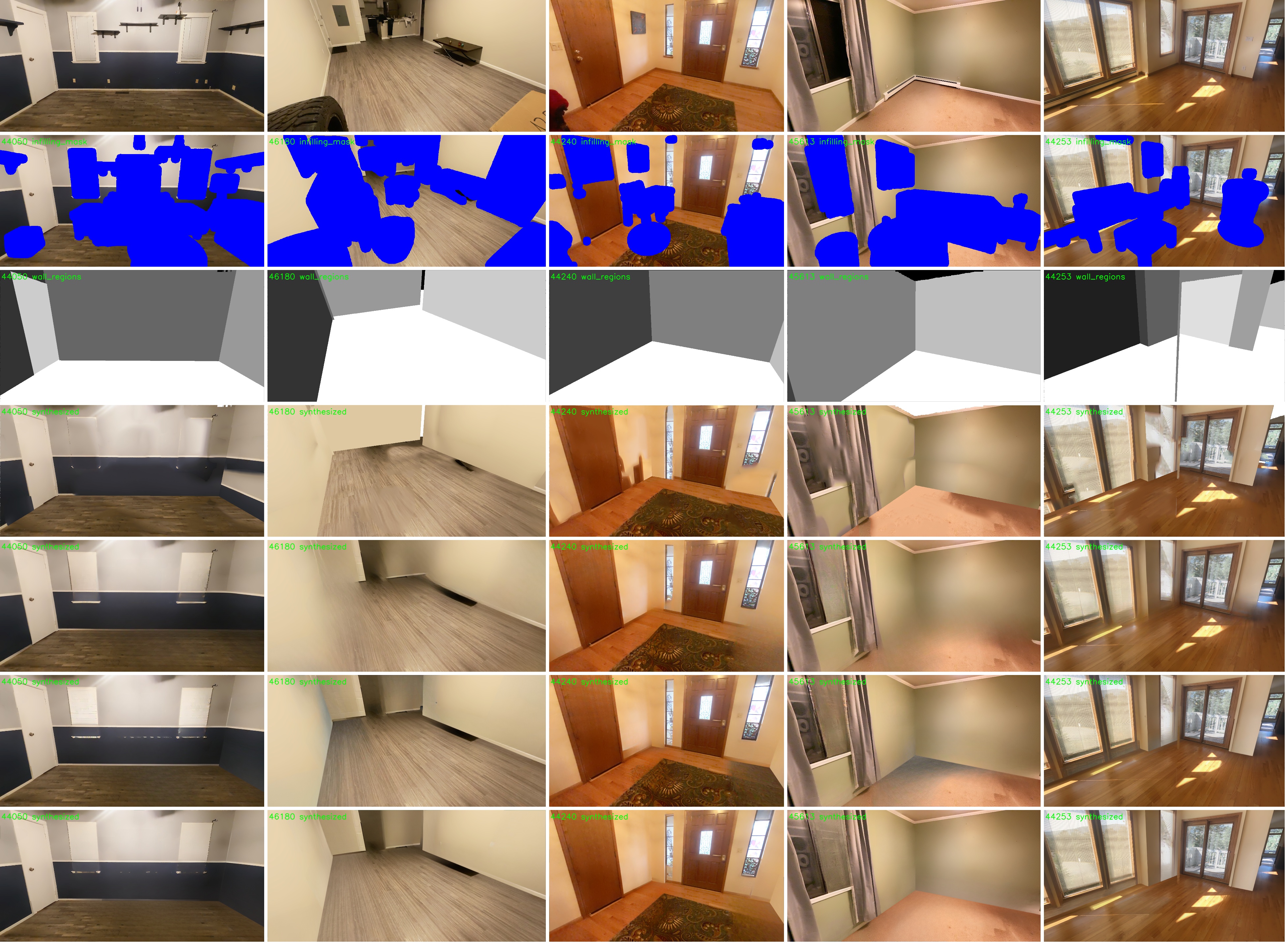}
    \caption{Qualitative examples from the test-set. Each column shows a single example over multiple methods. The top three rows are ground-truth image, masked input image, and the detected layout, respectively. From the fourth to seventh row, we show the prediction of: PatchMatch+P \cite{patchmatch}, LaMa\cite{lama}, LaMa+P, LaMa+P+R (ours). Abbreviations are described in \autoref{tab:results}.}
    \label{fig:abl}
\end{figure*}

We discuss the results of our evaluation, which are summarized in \autoref{tab:results}.
Starting from the top row, we report the performance of an implementation of PatchMatch \cite{patchmatch, patchmatch-jiayuan, patchmatch-younesse} applied in a per-plane manner.
Qualitatively, the inpainting isn't bad (row four of \autoref{fig:abl}), but there are many artifacts, like texture and structural discontinuities.
We next test LaMa \cite{lama} running directly on the input image without any modifications.
Numerically, it performs better than PatchMatch, but the inpainting has poor geometric consistency (row five of \autoref{fig:abl}).
We add the geometric constraint to \cite{lama} by performing inpainting using the Deep Neural Network, in a per-plane fashion.
It substantially improves the performance across all the metrics, and the results look much better (row six of \autoref{fig:abl}).
However, there is some blurry gray infill in some places, like the left wall in second example, or the floor in fourth example.
Our texture refinement step, shown in the last row in the figure, further improves the inpainting, and we are able to get rid of the gray areas completely.

\begin{table}[h!]
    \centering
    \begin{tabular}{c|ccc}
         & LPIPS$\downarrow$ & Incoherence$\downarrow$ & PSNR$\uparrow$ \\
        \hline
        PatchMatch\cite{patchmatch}+P & 0.4329 & 0.0203 & 17.001 \\
        LaMa\cite{lama} & 0.3715 & 0.0121 & 20.033\\
        LaMa+P & 0.3115 & 0.0063 & 20.689\\
        LaMa+P+R (ours) & \textbf{0.2858} & \textbf{0.0036} & \textbf{21.141} \\
    \end{tabular}
    \caption{Ablation study of the texture infill on indoor scenes from our in-house testset, using various methods. The abbreviations stand for: P-plane-wise infill, R-texture refinement} 
    \label{tab:results}
\end{table}

\section{Demonstration}

\begin{figure*}
    \centering
    \includegraphics[width=0.8\textwidth]{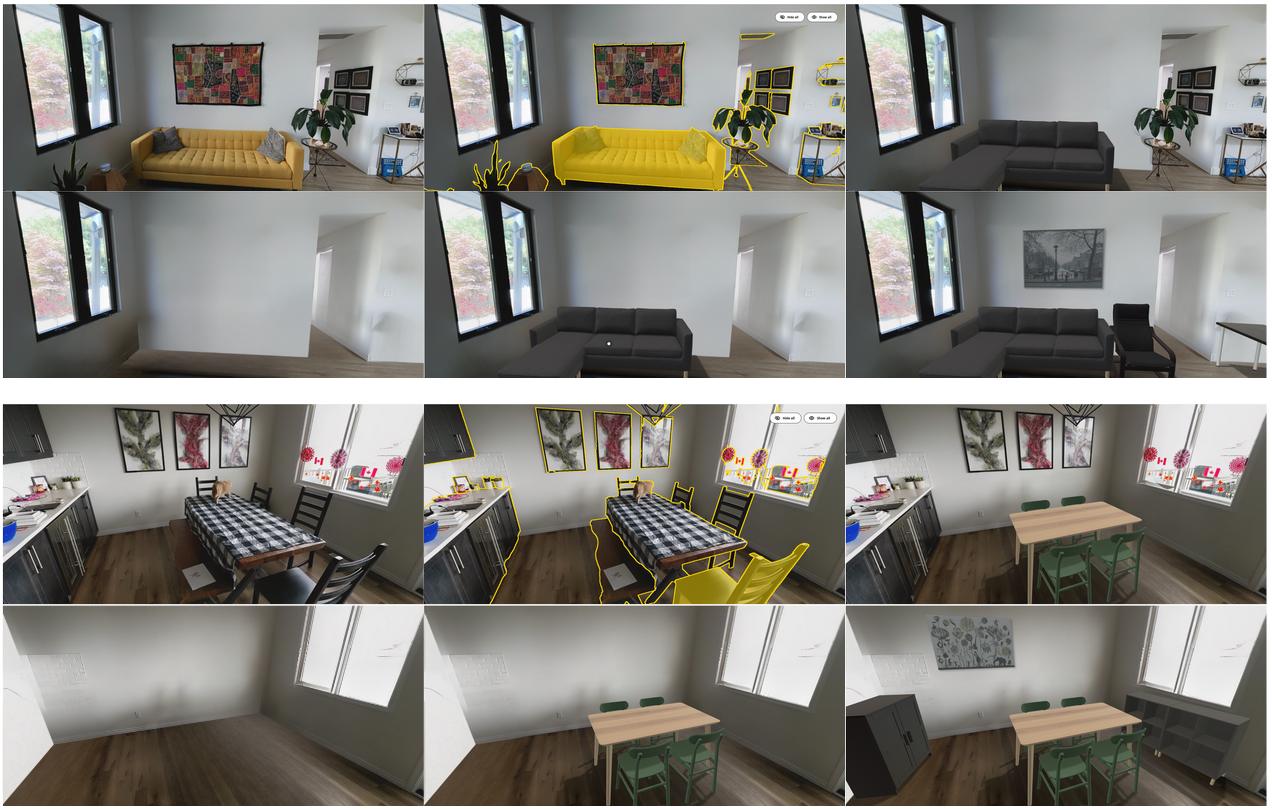}
    \caption{Example of a furniture eraser application on two scenes. For the top row, from left to right, we have input image, image showing furniture highlighted as selectables, and an image showing a real furniture replaced with a virtual one. The second row, from left to right, shows the room emptied using our method, emptied room with one virtual furniture, and emptied room with many  virtual furniture items. The next two rows are same sequence, but for another scene.} 
    \label{fig:edit}
\end{figure*}

We demonstrate how we can leverage our system to create a viable furniture eraser application.
The application starts with identifying all selectable objects with a highlighted outline.
Clicking on an object erases it from the scene and uses the inpainted depth so that the user can place virtual furniture in its place. Fig. \ref{fig:edit} shows two examples of rooms being edited by users.

\section{Conclusion}

We proposed a modular system for removing furniture from an indoor scene by leveraging perceptual cues such as segmentation and room layout. Our system performs per-plane inpainting using a deep inpainting network, and we show how removing any of the two components significantly degrades the inpainting performance. In the last section, we demonstrated the application of our system in re-decorating indoor spaces.

\bibliographystyle{abbrv-doi}

\bibliography{template}
\end{document}